\documentclass[journal]{IEEEtran}
\usepackage{amsmath,amsfonts}
\usepackage{algorithmic}
\usepackage{algorithm}
\usepackage{array}
\usepackage[caption=false,font=normalsize,labelfont=sf,textfont=sf]{subfig}
\usepackage{textcomp}
\usepackage{stfloats}
\usepackage{url}
\usepackage{verbatim}
\usepackage{graphicx}
\usepackage{cite}
\usepackage{booktabs}
\usepackage{xcolor}

\begin{document}

%\title{Simulation of Optical Tactile Sensors: High-fidelity, Scalable, and Dynamic Interaction Enabled}
\title{Simulation of Optical Tactile Sensors Supporting Slip and Rotation using Path Tracing and IMPM}
\author{Zirong Shen, Yuhao Sun, Shixin Zhang, Zixi Chen,~\IEEEmembership{Graduate Student Member, ~IEEE,} Heyi Sun\\ Fuchun Sun,~\IEEEmembership{Fellow,~IEEE,} and Bin Fang,~\IEEEmembership{Senior Member,~IEEE} 

% <-this % stops a space
\thanks{This work was supported by Major Project of the New Generation of Artificial Intelligence, China (No. 2018AAA0102900), the National Natural Science Foundation of China under Grant 62173197, U22B2042. Z. Shen and Y. Sun contributed equally to this work.  \emph{*Corresponding authors: Bin Fang.}} 
\thanks{Z. Shen and H. Sun are with the Zhili College, Tsinghua University, Beijing 100084, China. E-mail: {\tt\small shenzr21@mails.tsinghua.edu.cn; sun-hy21@mails.tsinghua.edu.cn}.}
\thanks{Y. Sun and B. Fang are with the school of Artificial Intelligence, Beijing University of Posts and Telecommunications, 100876, China. E-mail: {\tt\small yuhaosun0225@gmail.com;fangbin1120@bupt.edu.cn}.}
\thanks{F. Sun is with the Institute for Artificial Intelligence, Department of Computer Science and Technology, Beijing National Research Center for Information Science and Technology, Tsinghua University, Beijing 100084, China. E-mail: {\tt\small fcsun@mail.tsinghua.edu.cn}.}
\thanks{S. Zhang is with the School of Engineering and Technology, China University of Geosciences (Beijing), Beijing 100083, China. E-mail: {\tt\small zhangshixin@email.cugb.edu.cn}.}
\thanks{Z. Chen is with The BioRobotics Institute and the Department of Excellence in Robotics and AI, Scuola Superiore Sant’Anna, 56127 Pisa, Italy. E-mail: {\tt\small zixi.chen@santannapisa.it}.}
}
%询问这些信息具体是什么
 % \markboth{IEEE ROBOTICS AND AUTOMATION LETTERS,~Vol.~X, No.~X, XX~2022}%


\maketitle
\begin{abstract}
Optical tactile sensors are extensively utilized in intelligent robot manipulation due to their ability to acquire high-resolution tactile information at a lower cost. However, achieving adequate reality and versatility in simulating optical tactile sensors is challenging. 
In this paper, we propose a simulation method and validate its effectiveness through experiments. We utilize path tracing for image rendering, achieving higher similarity to real data than the baseline method in simulating pressing scenarios. Additionally, we apply the improved Material Point Method(IMPM) algorithm to simulate the relative rest between the object and the elastomer surface when the object is in motion, enabling more accurate simulation of complex manipulations such as slip and rotation.
% In this paper, we present an effective simulator that applied IMPM for objects moving on the sensor surface, coupled with path tracing for rendering, to address the complexities of light propagation. 
% %两个特点分开
% Experimental results demonstrate that our simulator achieves significantly higher similarity to real data than the baseline method in simulating static scenarios such as pressing and effectively supports the simulation requirements for dynamic scenarios like rotation and slip.
%dynamic -> complex manipulation
\end{abstract}

%2.23计划：先把8-9个图都画了，table也先放着

\section{Introduction}
%视触觉传感器-模拟
%[x] 可以直接从modern robots 开始

\IEEEPARstart{I}{ntelligent} robot manipulation plays a crucial role in human-robot interaction, hazardous environment exploration, and smart home applications.
Modern robots often exhibit high levels of autonomy to accomplish tasks such as recognizing objects\cite {19-Recognition} or grasping objects of various shapes and hardness\cite{Grasp-20}.
% [x] such as 后面两部分词性不一样
To autonomously decide each step of action, robots perceive the external environment through various types of sensors, such as temperature sensors, force sensors, and tactile sensors\cite{17-Sensor-Review}. Among various tactile sensors, optical tactile sensors stand out due to their advantages, such as high resolution and low cost.
% [x] 过渡：perceive - tactile -optical tactile
% \zc{one sentence for tactile sensors and next paragraph}

\begin{figure}
    \centering
    \includegraphics[width=3.4in]{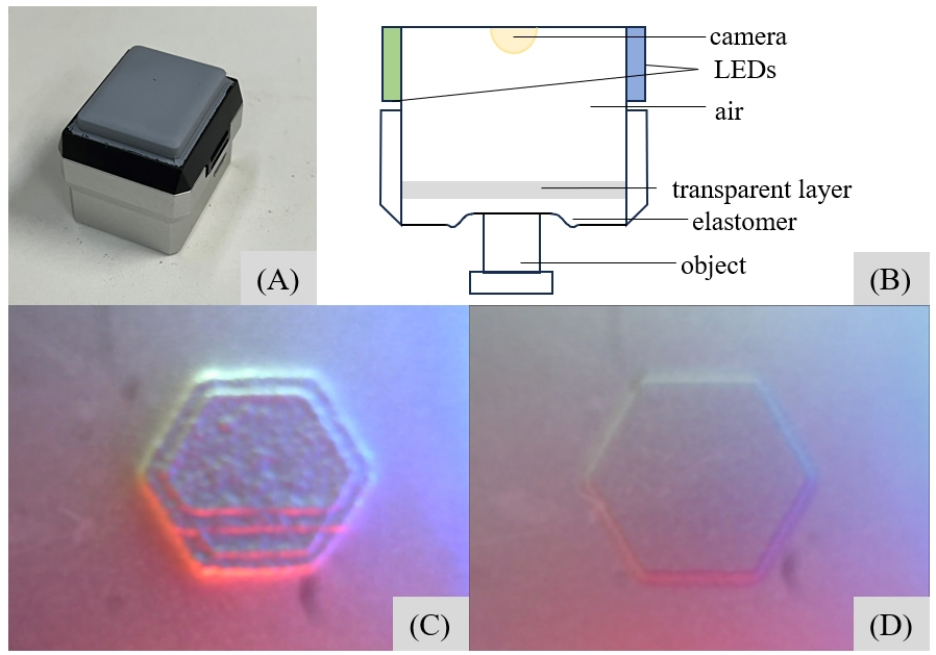}
    \caption{(A) The optical tactile sensor Gelsight. (B) Internal structure schematic of the Gelsight. (C) Real image captured by sensor camera. (D) Simulated image generated by our simulator. The images show that our simulator accurately simulates pressure boundaries, background textures, and lighting conditions.}
    \label{fig:gelsight}
\end{figure}

Recently, various optical tactile sensors have been developed, like Gelsight\cite{17-Gelsight} and 9DTact\cite{23-9DTact}. These sensors have found extensive applications in experiments related to shear and slip measurement\cite{15-Shear-slip,17-Slip}, hardness estimated\cite{16-hardness,17-hardness}, and so on. Most optical tactile sensors consist of a camera, a set of light sources, and an elastomer layer, as shown in Fig. \ref{fig:gelsight}-(\textbf{B}).
%\zc{as shown in Fig. x}. 
%Fig. \ref{fig:indenter}-(\textbf{A})
When an object approaches the sensor and comes into contact with the elastomer layer, the contact pressure causes the elastomer to deform, then the reflected light produced by the reflection of the reflective coating also changes accordingly. The camera can then capture the deformation of the elastomer, allowing tactile information to be captured from visual images. 
% [x] 从物体接触传感器-传感器的组成-硅胶变形-捕捉到反射光变化

% [x] 表示原因的词可以换一个（as）
Simulation plays a crucial role in robotics research, given that conducting real-world experiments consumes lots of time and resources and may lead to irreversible damage to hardware. Furthermore, reinforcement learning and neural networks are widely utilized in robotics like \cite{23-unigrasp, 18-grasp}. Simulation enables the generation of vast amounts of data that correspond to real-world scenarios in a shorter time and at lower costs, which is essential in generating datasets for neural networks and establishing reinforcement learning simulation environments.
%[x]重复了两个crucial/for，改一下
%[x]强化学习-dataset不是很合适；可以改成神经网络

%[x]第一句强调视触觉仿真重要，关键在于操作而不是神经网络
%\zc{decrease NN}
%超页数了，干脆把NN全删了
In the realm of research on optical tactile sensors, the application of simulation is significant. 
%Neural networks are also widely applied in the field of optical tactile sensors. As exemplified by \cite{18-grasp}, it applied a deep, multi-modal convolutional network. The network forecasts the success rates of various candidate grasping strategies at each step of the action and subsequently adapts to the most promising approach to execute the grasp. 
However, commercial robot simulators like Gazebo\cite{04-Gazebo}, MuJoCo\cite{12-MuJoCo}
% [x]like xxxx;单独使用商业的仿真软件对视触觉仿真困难；但有一些专门针对视触觉的仿真器
%可以去掉 单独
often struggle to simulate elastomer, because elastomer is elastic and provides complex deformations. Consequently, the acquisition of tactile data poses a substantial cost, representing a significant obstacle in related research endeavors.
% [x] 无法仿真elastomer应该是视触觉仿真中的问题
% 不要一直such as-like
The existing simulation methods for optical tactile sensors, including \cite{21-Gelsight-sim, 23-Tacchi,23-sim}, also face challenges like limited fidelity to real-world data and the incapacity to deal with fundamental operations like slip and rotation.
%[x] 不真实太宽泛了->对于常见的操作如旋转滑移支持不了
To address these issues, we propose a simulator that utilizes the Improved Material Point Method (IMPM) for simulation and ray tracing for rendering. 
The proposed simulator provides more realistic simulations of object pressing compared to previous methods. Additionally, it yields realistic results in simulating the slip and rotation of objects on the sensor surface.
%The proposed simulator yields realistic results in simulating the slip and rotation of objects on the sensor surface. Additionally, it provides more realistic simulations of object pressing compared to previous methods.
%press 相关的放在 complex 前面
%[x] 在MPM处强调可以支持旋转滑移
%[x] 改成-比之前的更好
The contributions of this paper are as follows:
\begin{enumerate}
\item We propose the IMPM algorithm, which improves the MPM to support a broader range of sensor operations, including slip and rotation. This enhancement enables the generation of high-quality simulation data.
%such as sliding and rotating, which can provide simulation datasets for tasks like detecting slipping and object grasping.
%[x]第二条只强调光线追踪就可以
%contribution 里不要放 未来能做的东西
\item We adopted the path tracing method to render simulation images, which can handle complex lighting conditions. Our method offers flexibility, transitioning from simulating a Gelsight sensor to the sensor shown in Fig. \ref{fig:thusensor}-(\textbf{A}) through simple modifications.
%Additionally, our method can easily adapted to various types of sensors through simple modifications.
%明确指出是两种传感器即可
%[x] 这里没必要有具体数据和效果-放到第三条
\item We validated the effectiveness of our simulation method through experiments. In press simulation, our method attains a Structural Similarity Index Measure (SSIM) similarity of 0.88 $\pm$ 0.05 between our simulation results and real-world data. In rotate and slip simulation, our improved method accurately simulates motion trace, aligning closely with real-world behavior.
%We conducted physical experiments and collected datasets of \todo{4} different objects under \todo{11} various distances of sliding and \todo{xx} different angles of rotation on the sensor. 

% [x]改成通过实验验证
\end{enumerate}

The rest of the paper is organized as follows: related works are reviewed in Sec. \ref{sec:2}; the method applied in elastomer simulation and rendering are described in Sec. \ref{sec:3}; the dataset in the experiments and the experimental procedure are introduced in Sec. \ref{sec:4}; experimental results are presented in Sec. \ref{sec:5}; the summary and outlook for the work are included in Sec. \ref{sec:6}.

\begin{figure*}
    \centering
    \includegraphics[width=7.1in]{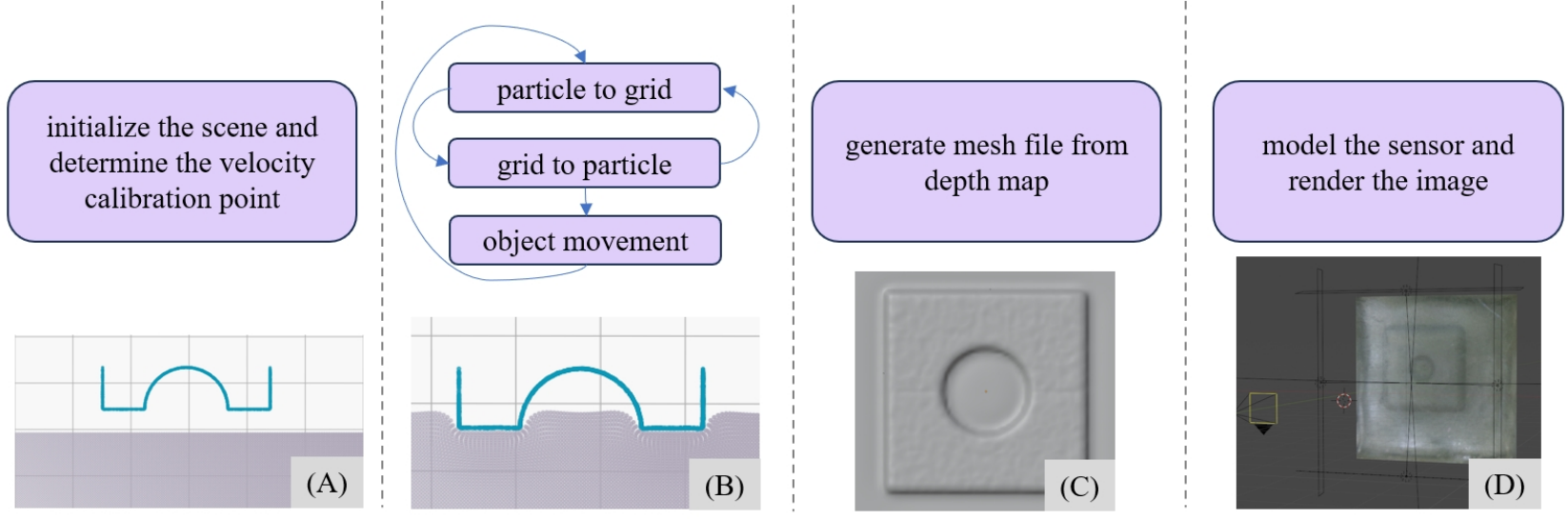}
    \caption{The entire simulation process pipeline. (A) Initialize the grids and particles in the MPM algorithm based on the object's initial position and angle. (B) Simulate the motions, such as pressing, sliding, and rotating step by step, until the target pose is achieved. (C) Interpolate the particle depths on the sensor surface to obtain a 3D elastomer model. (D) Add the lighting, materials, and texture effects to generate simulated images.}
    \label{fig:2}
    %[x] a,b,c,d的句子格式要一样
\end{figure*}

\section{Related Work}
\label{sec:2}
\subsection{Complex Operations with Optical Tactile Sensors}
\label{B.A}
In robotics research, various complex tasks, including object detection, grasping, and manipulation, require the assistance of tactile sensors. 
%[x]再加一句引入光学
Optical tactile sensors, with their advantages such as fine perception of the surroundings and high resolution, are well-suited for use in robots. By analyzing visual images and their variations, optical tactile sensors enable the acquisition of information about the shape, hardness, and movement of objects \cite{14-gel,16-hardness,17-hardness,17-Slip}.  
%或者把这一句放到下一段的开头
%Gelsight is one of the optical tactile sensors that meet the needs of diverse functions. It has a clear elastomer gel covered with a reflective membrane, and the elastomer gel is illuminated by four color LEDs.
%[x]多种传感器中，视触觉传感器的好处
%[x]需要一个视触觉传感器的例子，把传感器写的详细一点-Gelsight部分太重复了（和intro）

In \cite{14-gel}, the Fingertip Gelsight that can be mounted on the robot fingertip is proposed. The sensor is a cube with side lengths of 2.5 cm, featuring domed membranes on the surface and offering high resolution. By pressing the Fingertip GelSight against a known hemisphere, a mapping between color and depth is established. Subsequently, small objects can be localized and manipulated via visual images. 
%[x]在Gelsight Fingertip第一次出现的时候加引用
In \cite{16-hardness, 17-hardness}, researchers infer the contact force from the marker displacement and then estimate object hardness through the relationship between object deformation and contact force. The paper \cite{18-vitac, 19-cross} applies Gelsight to recognize cloth texture and employs the Deep Maximum Covariance Analysis (DMCA) and deep neural network to match the tactile features with visual features. Additionally, the authors of \cite{15-Shear-slip} place multiple markers on the elastomer surfaces and analyze the displacement of the markers to derive the displacement field. Afterward, they measure the shear, slip, and rotation of the object via the entropy of the shear displacement field. However, existing simulators often fail to simulate common operations like slip and rotation. 
%把slip那个引用换过来

To address these issues, we accounted for the relative rest of the indenter and elastomer surface in the simulation process and optimized specifically for slip and rotation.
%摩擦-> 考虑物体和硅胶表面的相对静止
%[x]视触觉传感器辅助复杂操作的方法（比如检测初始滑移（那篇应该又有滑移又有旋转），【x】通过按压来检测布料表面纹理等）
%可以用‘在上一个之外，这个xxx’表示并列
%复杂操作不好仿真-可以作为和A的连接点
%分段不要太多；时态统一%
%The ability to sense the position and movement of objects is one of the unique abilities of humans. Humans can determine whether objects are moving or stationary through visual and tactile. %不要太宽泛%
%In the field of robotics, object motion sensing is also important. 
%However, simulating optical tactile sensors is challenging because elastomer is a %compose of deformable material with high flexibility. 
%换个词，或者加点例子%
%One method of detecting movement is measuring the force between the robot and the object's contact surfaces. Multiple force sensors can be used for this method, such as Force Sensing Resistor\cite{} and Capacitive\cite{}.
 
% analyze 可以换成一个更具体的词
% slip and rotation 
% contribution: 1. deformation simulation 2. rendering 3. exp
% related work. 1. hard manipulation based on VBTS 2. VBTS simulation rendering
% 视触觉传感器的复杂操作/仿真的Rendering/复杂操作的仿真/实验验证，
% e.g.And chen et.al \cite{xxx} 可以用于标注作者
%And \cite{17-Slip} proposed that for objects with apparent features, movement of the object can be detected by tracking these features.
%使用一些有意义的连接词%

\subsection{Simulation of Optical Tactile Sensors}
\label{B.B}
%In recent years, various robot simulators have been developed, such as MuJoCo\cite{12-MuJoCo}, V-REP \cite{13-Vrep} and Gazebo\cite{04-Gazebo}.
In recent years, various simulation methods for optical tactile sensors have been proposed, including \cite{21-Gelsight-sim,22-Taxim,22-TACTO,23-Tacchi,24-difftactile}.

The authors of \cite{93-FEM, 19-FEM} utilize the Finite Element Method(FEM) to simulate the shape of elastomer when pressed. However, FEM requires significant computational resources for simulation and may lead to negative volumes in the simulated mesh when the object has significant deformation \cite{93-effect}. 
% apply, utilize, leverage, employ,  
In \cite{21-Gelsight-sim}, 2D Gaussian filters are employed to produce the elastomer height but only apply a smoothing process to the object shapes without considering specific deformation properties of the elastomer.
%conducting physically meaningful elastomer deformation simulations.
%弹性 deformation
%[x]没有物理性的仿真/只是smooth
Other simulators, including TACTO \cite{22-TACTO}, refine simulations using real-world sensor data and apply the Pyrender to render the images from synchronized scenes, while Taxim \cite{22-Taxim} employs the examples-based photometric stereo method and casts the shadow to render the images. Another approach presented in \cite{24-difftactile} combines FEM and MPM to simulate object-sensor interactions, and this work reconstructs the change of reflected color using a data-driven method. Meanwhile, the Tacchi \cite{23-Tacchi} employs the MLS-MPM method to simulate object-sensor interactions and the Phong model to render the images. Despite these advancements, accurately simulating the object and the sensor under complex manipulations remains a challenge that requires further research.
%Gelsight-sim，tacchi：Phong
%Tacto： Pyrender
%taxim
%difftactile
%对于以上传感器也加一下渲染方法
%[x] 不知道可否用‘引用的文章’当句子主语？ - 最好不要
%To address these issues前后的用法要呼应起来

%To address these issues, we propose a simulator based on Tacchi that offers more forms of movement, like rotation and slip.
%[x] 这个说法是有特定含义的，把特点写的更具体明确
To address these issues, we employ the IMPM to simulate the deformation of elastomer and the path tracing algorithm to render visual images from the depth map. Consequently, our simulator enables more accurate simulations of slip and rotation, better simulation of scenarios involving multiple reflections and refractions of light rays, and enhanced reality of simulated tactile images.
%[x] 或者是和真实图片的相似度增加
%[x] 添加光追的好处

\section{Method}
\label{sec:3}
In this section, our sensor simulator will be introduced. Subsection \ref{C.A} details the method of elastomer simulation. Following, Subsection \ref{C.B} introduces a tactile image rendering method based on ray tracing. 
%Finally, Subsection \ref{C.C} introduces the methods employed for evaluating the quality of the generated images.

\subsection{Elastomer Simulation}
\label{C.A}
%Material Point Method

This section describes the method applied in the first step of the simulation process, which involves simulating the contact between the elastomer and the object to obtain depth data on the elastomer surface after contact. The algorithm we utilized is an improved version of MPM (IMPM). MPM is an algorithm widely used in continuum materials simulation. Its advantages lie in its computational efficiency and its good support for large deformations. Therefore, it is used in many projects\cite{13-snow-mpm, 16-mpm}.

During the simulation process, the object and the elastomer are represented using particles, while the grid fills the entire simulation space, remaining stationary during motion. Each particle records the position $x\in R^3$, velocity $v\in R^3$, mass $m\in R$), affine velocity field $C\in R^{3\times 3}$ and deformation gradient $F\in R^{3\times 3}$. We divide continuous time into numerous time steps of length $\Delta t$ for simulation, during which the particle parameters can be considered invariant.
%In each step (现在which的指代不对)
Each simulation iteration can be broadly divided into five steps: particle-to-grid, handling grid boundary conditions, grid-to-particle, relative rest check, and particle movement.

\emph{particle to grid}:
Each particle transfers its information to nearby grid nodes. The mass of the $i$-th grid is
\begin{equation}
    M_i = \sum_{p\in \Omega_i}m_p w_{ip},
\end{equation}
where $\Omega_i$ is the particle in the $3\times 3\times 3$ grids around the i-th grid, $w_{ip}$ is the weight function between the $p$-th particle and the $i$-th grid, $m_p$ is the mass of the p-th particle. $w_{ip}$ is calculated based on the distances between particles and grid points in each dimension, as well as the cubic kernel function. Closer particles to the grid exert a stronger influence on it, with further details about $w_{ip}$ provided in \cite{23-Tacchi}.

The momentum of the grid $MG$ is calculated from the particle momentum $MM_i$ and elastic momentum $ME_i$, 
\begin{equation}
    MG_i=MM_i+ME_i.
\end{equation}
$MM_i$ and $ME_i$ are calculated from the nearby particle,
%\zc{more details, fewer eq. are better}
\begin{equation}
    MM_i=\sum_{p\in \Omega_i}w_{ip}(m_pv_p+C_p(X_g-x_p)),
\end{equation}
\begin{equation}
    ME_i=-\frac{4\Delta t}{W^2}\sum_{p\in \Omega_i}w_{ip}VI_pS_p(X_g-x_p),
\end{equation}
where $X_g$ represents the position of the grid where the $p$-th particle is located, $W$ represents the width of the grid, $VI_p$ represents the initial volume of the $p$-th particle, and $S_p$ represents the elasticity of the $p$-th particle\cite{21-elastic-sim}.

\emph{handling grid boundary conditions}:
Once the mass and momentum are obtained, the velocity of objects near the grid can be determined
\begin{equation}
    V_i=\frac{MG_i}{M_i}.
\end{equation}
%If a grid is located near the boundary and the velocity in a certain dimension points outward from the boundary, the velocity in that dimension needs to be set to zero.
Afterward, the velocities of the grid nodes at the boundaries are set to zero to prevent particles from leaving the simulation domain.

\emph{grid to particle}:
Once the grid velocities are obtained, the velocities of particles for the next time step can be updated. This process is similar to interpolating the velocities at the location of the particle from the velocities of the surrounding grids.
\begin{equation}
    v_p^{(k+1)}=\sum_{g\in G_p} w_{gp}V_g,
\end{equation}
\begin{equation}
    C_p^{(k+1)}=\frac{4}{W^2}\sum_{g\in G_p} w_{gp}v_p^{(k+1)}(X_i-x_p^{(k)}),
\end{equation}
\begin{equation}
    F_p^{(k+1)}=(I+\Delta tC_{p}^{(k+1)})F_{p}^{(k)},
\end{equation}
where $G_p$ is the $3\times 3\times 3$ grid around the $p$-th particle, $w_{gp}$ is the weight function between the $p$-th particle and the $g$-th grid.

\emph{relative rest check}:
%This section outlines our optimization approach for handling friction. - 是不是有点太啰嗦了……先删了
When the object moves parallel to the elastomer surface, such as in rotation and slip, the velocities of the object and the elastomer surface should closely match due to friction. When the object's velocity is set to a fixed value, multiple iterations of the particle-to-grid and grid-to-particle information exchange steps gradually align the elastomer velocity with that of the object. In this step, we initially set the velocities of all object particles to a specified value and set the z-axis velocity of the elastomer particles in the lower half layer to zero, as the bottom of the elastomer is fixed to the sensor. Then, we assess whether the velocities at the contact surface are sufficiently close. If so, we proceed with the subsequent steps; otherwise, we return to particle-to-grid and repeat the first three steps. 
%伪代码放上移动，这句话放到最后后面

\emph{particle movement}:
The velocity of all particles has been determined, allowing us to calculate their positions for the next time step,
\begin{equation}
    x_p^{(k+1)}=x_p^{(k)}+v_p^{(k+1)}\Delta t.
\end{equation}
The simulation process of IMPM within one time step $\Delta t$ is outlined in the Algorithm. \ref{algor:1}.

%Select two closely spaced elastomer particle and ojbect particle as representative points.
\begin{algorithm}
\caption{IMPM Algorithm}
\label{algor:1}
\begin{algorithmic}[1]
\STATE $x_{\text{o}} \leftarrow$ the bottom particle of the object
\STATE $x_{\text{e}} \leftarrow$ the closest elastomer particle to the point $x_{\text{o}}$.
\STATE flag $\leftarrow$ false
\STATE cnt $\leftarrow$ 0
\WHILE{flag = false}
\STATE Particle-to-grid
\STATE Handle boundary conditions
\STATE Grid-to-particle
\STATE Set the z-axis velocity of bottom elastomer particles to 0
\IF{$abs(\frac{v_{xe}-v_{xo}}{v_{xo}})\leq 
$ threshold}
%0.5->threshold value
\STATE Conduct a move step for a duration of $\Delta t$.
\STATE flag $\leftarrow$ true
\ENDIF
\STATE cnt $\leftarrow$ cnt + 1
\IF{cnt $\geq$ LIMIT}
\STATE flag $\leftarrow$ true
\ENDIF
\ENDWHILE
\STATE Particle movement
\end{algorithmic}
\end{algorithm}
\subsection{Path-tracing Method}
\label{C.B}

\begin{figure}
    \centering
    \includegraphics[width=3.4in]{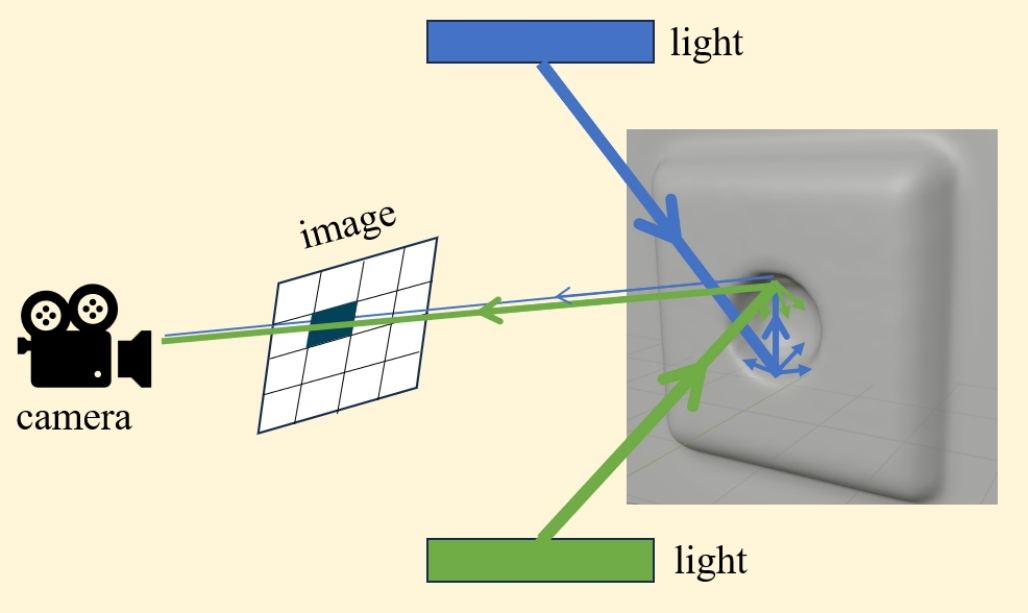}
    \caption{Light rays emitted from the blue LED undergo two reflections before reaching the camera, while those from the green LED undergo a single reflection to reach the camera. Upon combination, the resulting color at that point on the final image is the summation of the colors of the two rays. The line thickness represents light intensity.}
    \label{fig:ray-tracing}
\end{figure}

Path tracing simulates the path of light rays to model light transport, accounting for the color of each pixel in the resulting image, as shown in Fig. \ref{fig:ray-tracing}. Since light paths are reversible, we can emit rays from the camera at random angles, trace their trajectories, and calculate the directions of their reverse rays. By computing the attenuation and accumulation of colors along these paths, we can approximate the color of each pixel. 
%As the direction of light rays is arbitrary, increasing the number of randomly sampled angles can improve color fidelity to the true values for each pixel. 
In reality, light emitted from a source propagates in straight lines until it encounters a surface that obstructs its path, where it may be absorbed, refracted, or reflected. In the case of a semi-transparent medium, light can undergo partial refraction and reflection. Due to surface roughness, each reflection generates several diffuse rays in different directions. After each reflection, the intensity of the light ray diminishes.
%Consequently, a single ray may give rise to exponentially many rays, rendering it impractical to simulate all of them. Therefore, during the tracing process, it is crucial to establish a maximum number of bounces and a noise threshold to prevent infinite recursion. 
The advantage of using path tracing is its ability to handle scenarios where light rays reflect multiple times before reaching the camera (as illustrated in Fig. \ref{fig:ray-tracing}, the blue light undergoes two reflections before reaching the camera), resulting in high fidelity images. We applied the software Blender to model the scene and the physically-based path tracer in Blender - Cycles\cite{Cycles} for image rendering. Blender is a cross-platform and open-source 3D creation suite offering GPU rendering support, and it has wide-ranging applications, including \cite{17-Blender-1, 21-Blender-2,13-Blender-3}. 
%The advantage of employing ray tracing lies in its ability to simulate intricate optical properties of objects, such as the refraction effects caused by semi-transparent acrylic panels on the sensor surface and the reflection of light by baffles surrounding the sensor.
%[x]逐个/单独考虑每个像素，反向追踪光线
%Blender
%https://docs.blender.org/manual/en/latest/render/cycles/index.html
\begin{figure}
    \centering
    \includegraphics[width=3.4in]{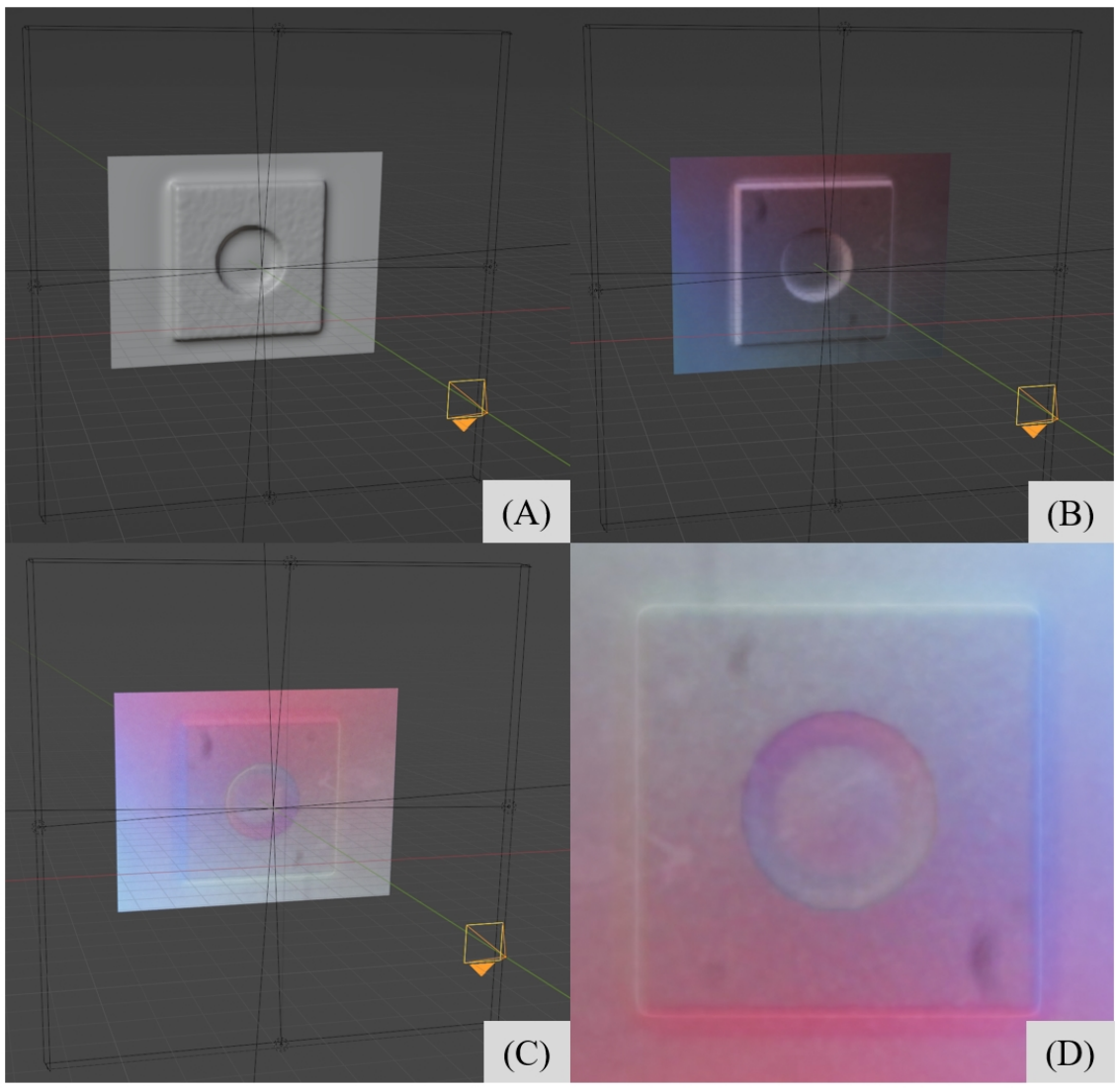}
    \caption{Illustration of the rendering process. (A) Depicts the bare elastomer model. (B) Superimposes a background image onto the sensor surface. (C) Incorporates LED lighting effects within the sensor. (D) Presents the resultant rendered image.}
    \label{fig:render}
\end{figure}

% \subsection{Evaluation Metric}
% \label{C.C}
% We assess the simulation effects by simulating the pressing conditions of Gelsight. The similarity between real and generated images can be applied to evaluate the extent to which the generated image matches reality. We utilized three metrics for evaluating the similarity: Mean squared error(MSE),  Peak Signal-to-Noise Ratio(PSNR), and  Structural Similarity(SSIM). These three metrics are commonly used for measuring image similarity \cite{22-Taxim, 22-TACTO, 23-Tacchi}.
% %[x] 引用一些使用了这些方法的论文
% MSE and PSNR calculate differences at each pixel between two images.
% %[x]m,n, i,j需要解释
% \begin{equation}
% MSE = \frac{1}{mn}\sum_{i=0}^{n-1}\sum_{j=0}^{m-1}[I(i,j)-I'(i,j)]^2,
% \end{equation}
% \begin{equation}
% PSNR = 10 \times \log_{10}(\frac{MAX^2}{MSE}),
% \end{equation}
% where I and I' represent the corresponding pixel values in the real and sim images, respectively, and m,n denote the dimensions of the images.

% SSIM considers local structural information of the images, measuring similarity using luminance, contrast, and structure, making it closer to human perception.
% \begin{equation}
% SSIM(x,y) = \frac{(2\mu_x\mu_y+c_1)(2\sigma_{xy}+c_2)}{(\mu_x^2+\mu_y^2+c_1)(\sigma_x^2+\sigma_y^2+c_2)},
% \end{equation}
% where $\mu$ is the pixel sample mean; $\sigma^2$ is the variance; $\sigma_{xy}$ is the covariance of x and y. And $c_1,c_2$ are variables to prevent the division by zero.

\section{Experimental Setup}
\label{sec:4}
    In this section, we perform a set of physical experiments alongside corresponding simulation experiments to evaluate the effectiveness of our simulation method. We conducted two types of experiments: press experiments were used to assess the optimization impact of ray tracing on light effects, while slip and rotation experiments were employed to evaluate the effectiveness of our elastomer deformation simulation. Subsection \ref{D.A} outlines the objects we used, along with the specific experimental methods employed. Subsection \ref{D.B} details how we simulated the sensors and the objects in the virtual environment.

\begin{figure}
    \centering
    \includegraphics[width=3.4in]{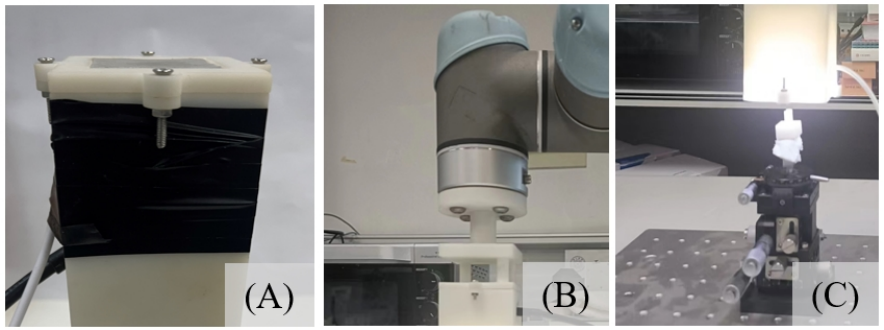}
    \caption{Slip and rotation data acquisition. (A) The sensor was utilized in the slip-and-rotate experiments. It consists of compact robotic skin, as detailed in \cite{24-thu-sensor}, along with white LED strips and a camera. (B) The connection between the robotic arm and the sensor. The robotic arm is employed to regulate the initial position and press depth of the sensor. (c) The control device of the object provides precise movements at specified distances and angles.}
    %这样变成（A）是主语了，改成后面是完整句子 或者\subgraph(A)xxx
    \label{fig:thusensor}
\end{figure}

\subsection{Real World Setup}
\label{D.A}
\begin{figure}
    \centering
    \includegraphics[width=3in]{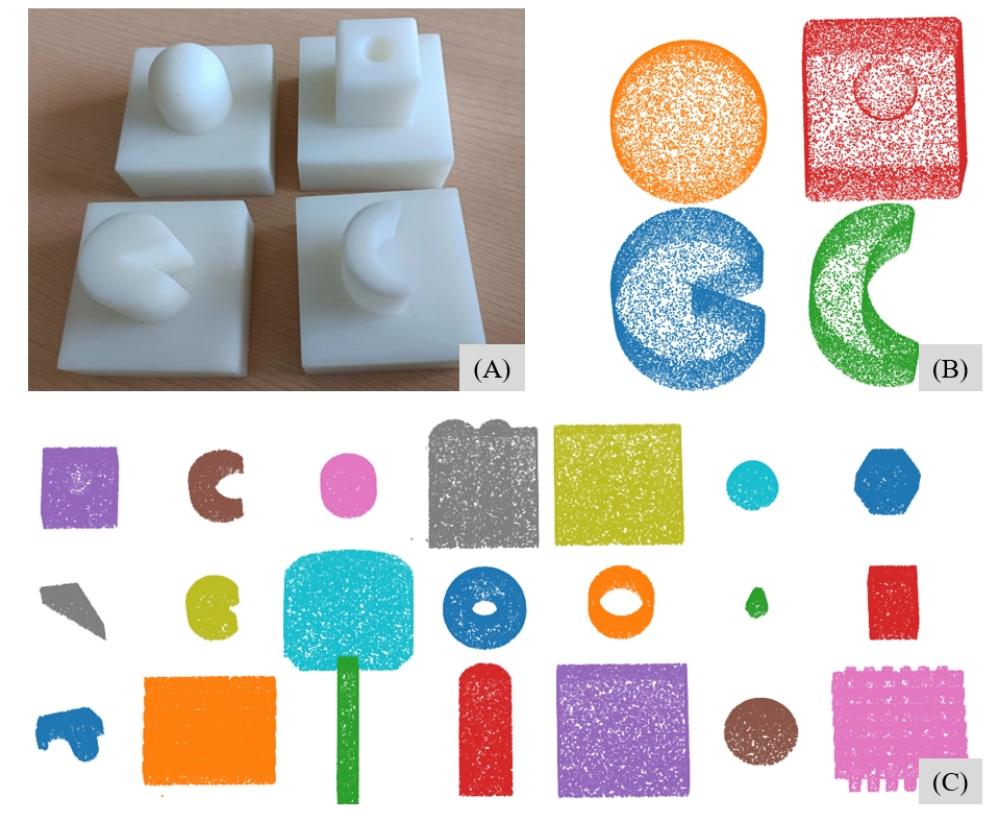}
    \caption{(A) Represents the indenter utilized in physical slip and rotation experiments. (B) Depicts the indenter employed in simulated rotation and pressing experiments. (C) Illustrates the indenter used in simulated pressing experiments, with detailed information available in \cite{21-Gelsight-sim}.}
    \label{fig:indenter}
\end{figure}
In the motion experiments, we employed four types of indenters: moon, pacman, dot\_in, and sphere
%长破折号似乎是不存在的
(as illustrated in Fig. \ref{fig:indenter}-(\textbf{A}) ). These indenters underwent deliberate rounding of their edges, prompted by our preliminary findings indicating that sharp edges could potentially damage the coating on the sensor surface. We utilized the sensor depicted in Fig.\ref{fig:thusensor}-(\textbf{A}) to collect data because it features rupture resistance, enabling it to withstand ample deformation in experiments, with noticeable traces left behind after movement. Initially, we connect the sensor and the robotic arm(as shown in Fig. \ref{fig:thusensor}-(\textbf{B})), securing the indenter onto the mobile platform(as shown in Fig. \ref{fig:thusensor}-(\textbf{C})). We then manipulate the robotic arm to align the center of the sensor with the object and vertically press to a depth of 0.5mm. Subsequently, precise movement of the indenter is achieved by operating the mobile platform. To assess the simulator's performance across diverse scenarios, we collected data on four types of indenters sliding leftward and rightward, all starting from the midpoint. Each direction covered a total distance of 5mm, with increments of 1mm, resulting in a total of $4\times 2 \times 6 = 48$ images. Concerning rotation, we collected data on three indenter types (excluding the sphere, as it undergoes minimal changes during rotation) rotating clockwise and counterclockwise. The rotation angles ranged from 0 to 45 degrees, with increments of 5 degrees in each direction, resulting in a total of $3\times 2 \times 10 = 60$ images.

In the press experiment, we utilized the dataset from \cite{21-Gelsight-sim}, which was gathered using Gelsight. This dataset comprises 21 indenters, each pressed at $3\times 3$ locations with depths ranging from 0 to 10 mm at 1 mm intervals, resulting in a total of $21\times 9 \times 11 = 2079$ images.
%[x]引用罗老师以前的论文-方法

\subsection{Virtual World Setup}
\label{D.B}
The entire simulation process is illustrated in Fig. \ref{fig:2}. The first step involves initializing the elastomer and the object. As shown in Fig. \ref{fig:2}-(\textbf{A}), the blue particles represent the object, and the gray particles represent the elastomer. The elastomer layer is initialized as a cuboid consisting of $201\times 201\times 41$ particles, while the object is initialized with its center point aligned with the center of the elastomer. Subsequently, velocity calibration points are identified, with the lowest point (closest to the elastomer) of the object designated as $x_o$ and the nearest elastomer particle to $x_o$ designated as $x_e$(as shown in Algorithm. \ref{algor:1}). 
%In the move simulation, we set Young's modulus to $1.5 \times 10^5$ and Poisson's ratio to $0.4$. This choice is based on preliminary experiments, where these physical parameters yielded simulated traces that closely matched those observed during the sliding of the actual object. 
During the simulation process, we utilize the positional changes of the previously selected calibration points $x_o$ to measure the distance and angle of object movement. This method is adopted because the velocities of the object and elastomer influence each other. Using this method for measurement is more accurate than direct calculation using speed and time.

After the deformation simulation, we extract the velocity of the elastomer surface particle, as the images captured by the optical tactile sensor are solely influenced by the light and the shape of the elastomer surface. Subsequently, we introduce minor perturbations to the depth map.
%Subsequently, we apply a certain amount of noise to the depth map, following a normal distribution $N(0,0.0001)$. 
%不要写详细的且无法解释的数
This prevents depth sameness in the unpressed areas, which could otherwise lead to erroneous normal vector directions on some faces in the generated mesh file, potentially resulting in textures being wrapped to the wrong side of the mesh. Afterward, we utilize libraries such as interpolate and open3d to convert the depth image into a mesh file with dimensions of $L\times W$ ($640\times 480$ for the Gelsight and $480\times 480$ for the sensor utilized in slip and rotation experiments).
%We applied the software Blender to model the scene and the physically-based path tracer in Blender - Cycles\cite{Cycles} for image rendering. Blender is a cross-platform and open-source 3D creation suite offering GPU rendering support, and it has wide-ranging applications, including \cite{17-Blender-1, 21-Blender-2,13-Blender-3}. 
We employed a Python script to automate modeling and rendering using the interfaces provided by Blender. Firstly, we import the mesh file of the previously generated elastomer layer, as shown in Fig. \ref{fig:render}-(\textbf{A}). Next, we wrap the base texture of the sensor (images captured when not in contact with objects) onto the object using UV mapping, as depicted in Fig. \ref{fig:render}-(\textbf{B}). Subsequently, lighting effects are added, as illustrated in Fig. \ref{fig:render}-(\textbf{C}).  We set the roughness of the elastomer to its maximum value, resulting in predominantly diffuse reflections, which closely resemble the actual elastomer layer. For lighting, we optimized the RGB values of the four-color LED in Gelsight simulation by sampling colors from real images. In the slip and rotation experiments, all lights emit white light, so each light has an RGB value of (255,255,255) Finally, with a path tracing sample count set to 128, we render the visual image, as seen in Fig. \ref{fig:render}-(\textbf{D}).
% \begin{table}[!ht]
% \caption{RGB values of the LED in Gelsight simulation}
% \centering
% \begin{tabular}{l|c c}
% % \hline
% \toprule
% & (R, G, B)&energy\\
% \midrule
% White & (255, 255, 255)&$10^3$\\
% Green & (200, 255, 200)&$2\times 10^3$\\
% Blue &(25, 75, 255)&$2\times 10^3$\\
% Red &(255, 50, 75)&$2\times 10^3$\\
% \bottomrule
% \end{tabular}
% \label{table:color}
% \end{table}

%Then, this file was imported into the scene, and it serves as the elastomer surface in the sensor. Meanwhile, four surface light sources of different RGB colors were placed around the model in accordance with the Gelsight style. Subsequently, the background image was applied as a texture to this surface, with the roughness adjusted to its maximum to approximate the optical properties of the real surface. Finally, the simulation images were obtained by the Cycles engine with 128 iterations of ray tracing.

\section{Experimental Results}
\label{sec:5}
This section showcases the simulation experiment results of our method. Subsection \ref{E.A} demonstrates the optimization of path tracing for image rendering. Subsection \ref{E.B} illustrates the effect of the IMPM in motion simulation.
\subsection{Press Simulation}
\label{E.A}
\begin{figure*}
    \centering
    \includegraphics[width=7.1in]{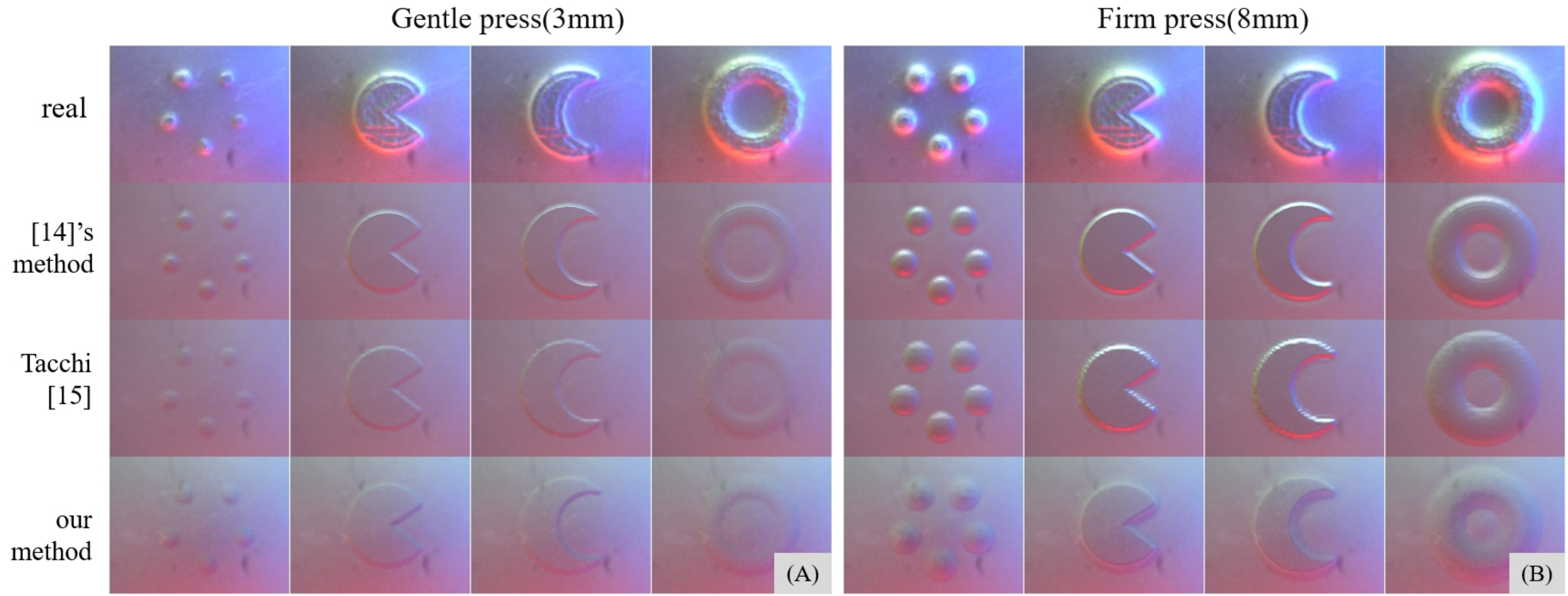}
    \caption{Comparison of Gelsight simulation results. (A) and (B) illustrate the cases with pressing depths of 3mm and 8mm, respectively. In each picture, the first row displays images captured by a real Gelsight sensor. The second row showcases simulation results obtained using the Difference of Gaussians proposed in \cite{21-Gelsight-sim}. The third row presents simulation results obtained by Tacchi \cite{23-Tacchi}. The fourth row displays results obtained using our simulation method. Four types of indenters have been selected for pressing demonstrations: dots, pacman, moon, and torus, showcasing rendering scenarios of curved edges, straight edges, and complex edges, respectively. In the simulated images generated by our method, the background appears clearer.}
    \label{fig:result-gelsight}
\end{figure*}
%[x]写上背景更清晰

We deploy the results from prior work \cite{21-Gelsight-sim,23-Tacchi} as benchmark reference. The DoG \cite{21-Gelsight-sim} applies the Difference of Gaussians to simulate the Gelsight, while the Tacchi \cite{23-Tacchi} applies the MLS-MPM to simulate the deformation of elastomer. Both of them employ the Phong model to render the image. 
%Due to a global offset between the positions of the indenters in the three simulation methods and the real dataset, we utilized the dot\_in indenter with clearly defined boundaries as an alignment standard. 
%概述一下有对齐就行了
%We then cropped the real images to $[0,590]\times[30,480]$ and the simulated images to $[25,615]\times[0,450]$ to as the comparative region. 
Due to a global offset between the positions of the indenters in the three simulation methods and the real dataset, we performed global cropping and alignment on the images before comparison. Fig. \ref{fig:result-gelsight} presents the real images alongside the simulated images generated using the DoG, Tacchi, and our method (aligned after cropping).  We utilized three metrics for evaluating the similarity: Mean squared error(MSE),  Peak signal-to-noise ratio (PSNR), and  Structural Similarity(SSIM). These three metrics are commonly used for measuring image similarity \cite{22-Taxim, 22-TACTO, 23-Tacchi}. In the press simulation, Tacchi and our method approach for elastomer deformation are fundamentally similar, with the main distinction being that Tacchi employs the Phong model while we utilize path tracing. Table \ref{table1} presents the results of evaluating and averaging 2079 sets of data. Our method outperforms previous methods in all three metrics.

\begin{table}[!ht]
\caption{image quality comparison of render method}
\centering
\begin{tabular}{l|c c c}
% \hline
\toprule
& PSNR $\uparrow$ & SSIM $\uparrow$& MSE $\downarrow$ \\
\midrule
%\cite{21-Gelsight-sim}'s method           & {$17.94\pm1.7$}    & {$0.84\pm0.07$} & {$1129.27\pm466.77$}\\
Phong \cite{23-Tacchi}          & {$18.1\pm1.62$}    & {$0.85\pm0.07$} & {$1079.73\pm416.3$}\\
Path tracing & {$\mathbf{19.94\pm1.86}$}    & {$\mathbf{0.88\pm0.05}$} & {$\mathbf{717.9\pm277.83}$}\\
\bottomrule
\end{tabular}
\label{table1}
\end{table}

\begin{figure}[!ht]
    \centering
    \includegraphics[width=3.4in]{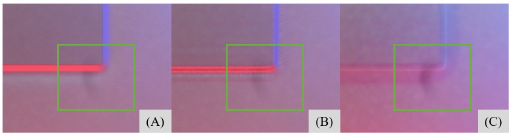}
    \caption{Right bottom corner of the image under the 8mm press of the prism indenter. The three images (A), (B), and (C) depict the results obtained using the DoG, Tacchi, and our method, respectively. In the first two images, the dark features within the green box remain unchanged in position and are barely noticeable. In our method, these features are visibly stretched due to the pressing action of the object.}
    \label{fig:detail}
\end{figure}
Regarding image details, the left side of the real image is illuminated in blue, while the bottom is red, with a smooth transition at the pressing edge. In our method, the influence of lighting on the background is consistent with reality, whereas in the Phong model, the unpressed areas are almost uniformly colored and not illuminated. When the pressing depth is deep, %a depression appears one circle outside the edge of DoG, while 
Tacchi's edge shows a sawtooth, while our method performs well under various pressing depths. Since our method wraps the background image as a texture onto the elastomer, the colors and features in the background are better preserved. Moreover, when the features in the background coincide with the pressed region, our method can present the effect of features following depth changes. In contrast, the background features in the other two methods remain in their original positions. Fig. \ref{fig:detail} presents a magnified section of the simulated images, highlighting the above characteristics. This advantage could be exploited in the future for simulating sensors with marked points.

\subsection{Complex Manipulation Simulation}
\label{E.B}

\begin{figure}
    \centering
    \includegraphics[width=3.4in]{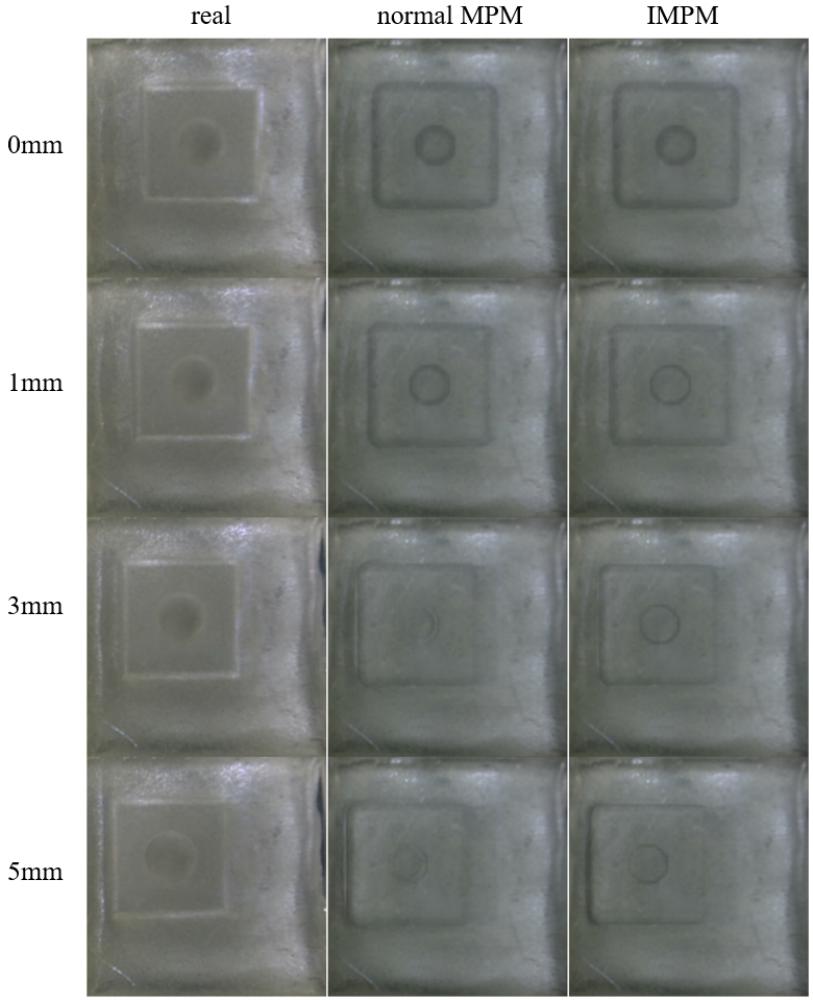}
    \caption{Comparison of Slip Simulation Results. The first, second, and third columns represent real data, simulation using the original MPM, and simulation using the IMPM, respectively. From top to bottom, the scenarios depict the dot\_in indenter pressing and then sliding leftward by 0mm, 1mm, 3mm, and 5mm.}
    \label{fig:6}
\end{figure}
\begin{figure}[!ht]
    \centering
    \includegraphics[width=3.4in]{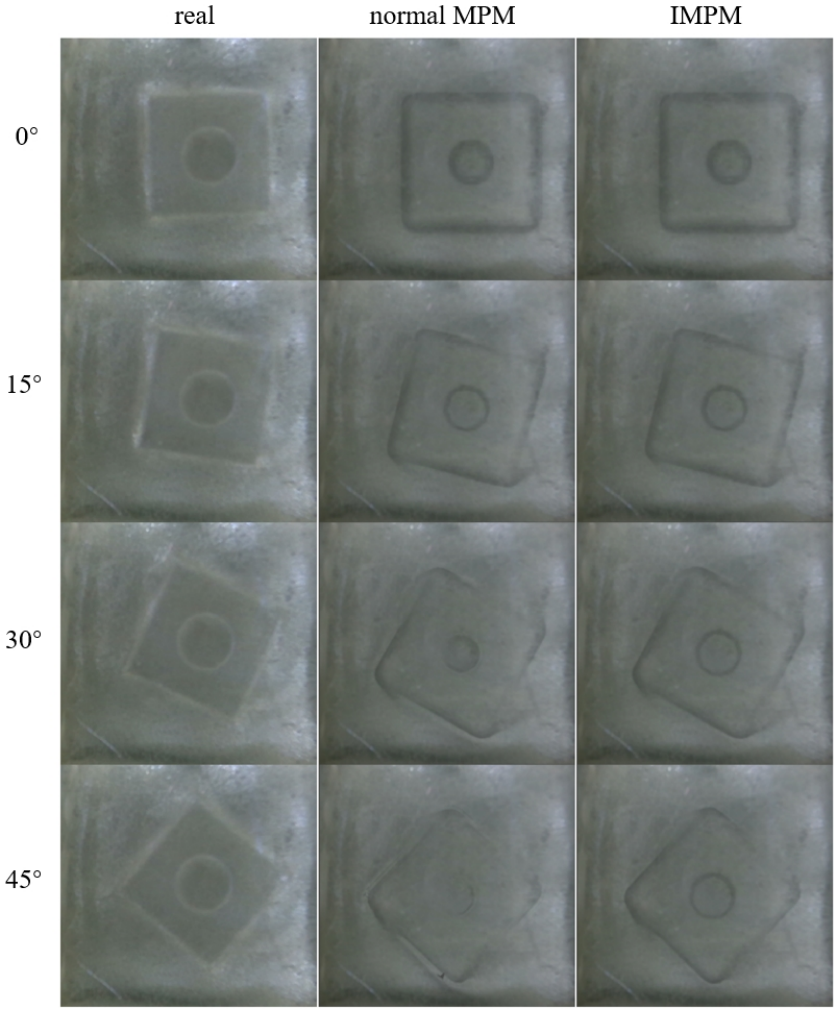}
    \caption{Comparison of Rotation Simulation Results. Progressing from top to bottom, the scenarios depict the dot\_in indenter pressing and then rotating clockwise by 0°, 15°, 30°, and 45°.}
    \label{fig:7}
\end{figure}

In this section, we evaluate the impact of the IMPM on slip and rotation simulation. The "normal" group utilizes the original MPM without any enhancements, while the IMPM group employs the algorithm proposed in Section \ref{C.A}. Fig. \ref{fig:6} illustrates the scenarios of real images, simulation images with the normal MPM, and simulation images with the IMPM for dot\_in indenter sliding leftward by 0mm, 1mm, 3mm, and 5mm, while Fig. \ref{fig:7} depicts the rotation of the dot\_in indenter clockwise by 0, 15, 30, and 45 degrees. Table \ref{table:slip} evaluates the performance of both simulation methods, indicating improvements in all three parameters after the enhancement.

In terms of slip simulation, it is observed that as the sliding progresses, some regions transition from being pressed to unpressed, while some exhibit the opposite behavior. In real images, areas no longer pressed transform into dragging traces, characterized by shallow indentations, yet the overall shape of the pressed area remains largely unchanged. In the normal MPM, once an area is no longer pressed, significant depth variety occurs, with almost no visible central hole in the indenter at the image of 5mm. In contrast, the IMPM mitigates this issue. 

\begin{table}[!ht]
\caption{image quality comparison of slip}
\centering
\begin{tabular}{l|c c c}
% \hline
\toprule
& PSNR $\uparrow$ & SSIM $\uparrow$& MSE $\downarrow$ \\
\midrule
normal & {$23.09 \pm 0.24$} & {$0.856 \pm 0.006$} & {$319.4 \pm 17.49$} \\
IMPM  & {$\mathbf{23.22 \pm 0.27}$} & {$\mathbf{0.857 \pm 0.006}$} & {$\mathbf{310.56 \pm 18.94}$} \\
\bottomrule
\end{tabular}
\label{table:slip}
\end{table}

In terms of rotation simulation, the normal MPM exhibits another issue: the circular hole portion remains unpressed throughout the rotation, so the depth should not vary significantly. However, due to the lack of consideration for fixing the bottom of the elastomer, it is subjected to a downward force at a slant angle after rotating for some time, leading to a deeper but wrong indentation. Additionally, with the inclusion of frictional force simulation, the rotational traces generated by the IMPM are closer to reality. Table \ref{table:rotate} assesses the effectiveness of both simulation methods, showing only marginal improvement with the IMPM over the normal MPM. This is because, during the rotation process, only small areas along the edges of the probe are affected, resulting in minor differences in overall image similarity.

\begin{table}[!ht]
\caption{image quality comparison of rotation}
\centering
\begin{tabular}{l|c c c}
% \hline
\toprule
& PSNR $\uparrow$ & SSIM $\uparrow$& MSE $\downarrow$ \\
\midrule
normal& {$26.51\pm 0.77$} & {$0.89\pm 0.006$} & {$147.43\pm 26.73$} \\
IMPM & {$\mathbf{26.56\pm 0.81}$} & {$0.89\pm 0.006$} & {$\mathbf{146.27\pm 27.84}$} \\
\bottomrule
\end{tabular}
\label{table:rotate}
\end{table}

\section{Conclusion and Discussion}
\label{sec:6}
%This paper proposes a simulation method for optical tactile sensors, with targeted optimizations for the friction between dynamic objects and sensors, coupled with modeling of the sensor and employing path tracing algorithms to generate images. 
This paper presents a simulation method for optical tactile sensors. It employs the path tracing algorithm to simulate the sensor and generate simulation images and proposes the IMPM algorithm to address the relative rest between the object and the elastomer surface during slip and rotation.

The method exhibits high scalability, accommodating various sensors. For some sensors like \cite{17-Slip}, where the elastomer layer is not cuboid-shaped, our method can adapt to them by adjusting the shape of the elastomer particle cloud during deformation simulation. Similarly, adjustments can be made during modeling to accommodate different reflective layers, lighting conditions, and sensor shell shapes.

However, the method faces certain challenges, including suboptimal simulation efficiency. For instance, on a laptop equipped with an NVIDIA GeForce RTX 3060 GPU, path tracing takes approximately 3.8 seconds per frame. Another issue is that we cannot accurately simulate the relative sliding between the object and the sensor surface during long-distance movement. Future research will focus on refining the handling of contact between objects.
%Additionally, the simulation only considers static friction as the primary frictional force between objects and sensors without adapting to different material properties of objects. 
\bibliographystyle{IEEEtran}
\bibliography{IEEEabrv,references}

\begin{thebibliography}{10}
\providecommand{\url}[1]{#1}
\csname url@samestyle\endcsname
\providecommand{\newblock}{\relax}
\providecommand{\bibinfo}[2]{#2}
\providecommand{\BIBentrySTDinterwordspacing}{\spaceskip=0pt\relax}
\providecommand{\BIBentryALTinterwordstretchfactor}{4}
\providecommand{\BIBentryALTinterwordspacing}{\spaceskip=\fontdimen2\font plus
\BIBentryALTinterwordstretchfactor\fontdimen3\font minus \fontdimen4\font\relax}
\providecommand{\BIBforeignlanguage}[2]{{%
\expandafter\ifx\csname l@#1\endcsname\relax
\typeout{** WARNING: IEEEtran.bst: No hyphenation pattern has been}%
\typeout{** loaded for the language `#1'. Using the pattern for}%
\typeout{** the default language instead.}%
\else
\language=\csname l@#1\endcsname
\fi
#2}}
\providecommand{\BIBdecl}{\relax}
\BIBdecl

\bibitem{19-Recognition}
J.~M. Gandarias, A.~J. García-Cerezo, and J.~M. Gómez-de Gabriel, ``Cnn-based methods for object recognition with high-resolution tactile sensors,'' \emph{IEEE Sensors Journal}, vol.~19, no.~16, pp. 6872--6882, 2019.

\bibitem{Grasp-20}
C.~Gabellieri, F.~Angelini, V.~Arapi, A.~Palleschi, M.~G. Catalano, G.~Grioli, L.~Pallottino, A.~Bicchi, M.~Bianchi, and M.~Garabini, ``Grasp it like a pro: Grasp of unknown objects with robotic hands based on skilled human expertise,'' \emph{IEEE Robotics and Automation Letters}, vol.~5, no.~2, pp. 2808--2815, 2020.

\bibitem{17-Sensor-Review}
\BIBentryALTinterwordspacing
L.~Zou, C.~Ge, Z.~J. Wang, E.~Cretu, and X.~Li, ``Novel tactile sensor technology and smart tactile sensing systems: A review,'' \emph{Sensors}, vol.~17, no.~11, 2017. [Online]. Available: \url{https://www.mdpi.com/1424-8220/17/11/2653}
\BIBentrySTDinterwordspacing

\bibitem{17-Gelsight}
\BIBentryALTinterwordspacing
W.~Yuan, S.~Dong, and E.~H. Adelson, ``Gelsight: High-resolution robot tactile sensors for estimating geometry and force,'' \emph{Sensors}, vol.~17, no.~12, 2017. [Online]. Available: \url{https://www.mdpi.com/1424-8220/17/12/2762}
\BIBentrySTDinterwordspacing

\bibitem{23-9DTact}
C.~Lin, H.~Zhang, J.~Xu, L.~Wu, and H.~Xu, ``9dtact: A compact vision-based tactile sensor for accurate 3d shape reconstruction and generalizable 6d force estimation,'' 2023.

\bibitem{15-Shear-slip}
W.~Yuan, R.~Li, M.~A. Srinivasan, and E.~H. Adelson, ``Measurement of shear and slip with a gelsight tactile sensor,'' in \emph{2015 IEEE International Conference on Robotics and Automation (ICRA)}, 2015, pp. 304--311.

\bibitem{17-Slip}
S.~Dong, W.~Yuan, and E.~H. Adelson, ``Improved gelsight tactile sensor for measuring geometry and slip,'' in \emph{2017 IEEE/RSJ International Conference on Intelligent Robots and Systems (IROS)}, 2017, pp. 137--144.

\bibitem{16-hardness}
W.~Yuan, M.~A. Srinivasan, and E.~H. Adelson, ``Estimating object hardness with a gelsight touch sensor,'' in \emph{2016 IEEE/RSJ International Conference on Intelligent Robots and Systems (IROS)}, 2016, pp. 208--215.

\bibitem{17-hardness}
W.~Yuan, C.~Zhu, A.~Owens, M.~A. Srinivasan, and E.~H. Adelson, ``Shape-independent hardness estimation using deep learning and a gelsight tactile sensor,'' in \emph{2017 IEEE International Conference on Robotics and Automation (ICRA)}, 2017, pp. 951--958.

\bibitem{23-unigrasp}
W.~Wan, H.~Geng, Y.~Liu, Z.~Shan, Y.~Yang, L.~Yi, and H.~Wang, ``Unidexgrasp++: Improving dexterous grasping policy learning via geometry-aware curriculum and iterative generalist-specialist learning,'' 2023.

\bibitem{18-grasp}
R.~Calandra, A.~Owens, D.~Jayaraman, J.~Lin, W.~Yuan, J.~Malik, E.~H. Adelson, and S.~Levine, ``More than a feeling: Learning to grasp and regrasp using vision and touch,'' \emph{IEEE Robotics and Automation Letters}, vol.~3, no.~4, pp. 3300--3307, 2018.

\bibitem{04-Gazebo}
N.~Koenig and A.~Howard, ``Design and use paradigms for gazebo, an open-source multi-robot simulator,'' in \emph{2004 IEEE/RSJ International Conference on Intelligent Robots and Systems (IROS) (IEEE Cat. No.04CH37566)}, vol.~3, 2004, pp. 2149--2154 vol.3.

\bibitem{12-MuJoCo}
E.~Todorov, T.~Erez, and Y.~Tassa, ``Mujoco: A physics engine for model-based control,'' in \emph{2012 IEEE/RSJ International Conference on Intelligent Robots and Systems}, 2012, pp. 5026--5033.

\bibitem{21-Gelsight-sim}
D.~F. Gomes, P.~Paoletti, and S.~Luo, ``Generation of gelsight tactile images for sim2real learning,'' \emph{IEEE Robotics and Automation Letters}, vol.~6, no.~2, pp. 4177--4184, 2021.

\bibitem{23-Tacchi}
Z.~Chen, S.~Zhang, S.~Luo, F.~Sun, and B.~Fang, ``Tacchi: A pluggable and low computational cost elastomer deformation simulator for optical tactile sensors,'' \emph{IEEE Robotics and Automation Letters}, vol.~8, no.~3, pp. 1239--1246, 2023.

\bibitem{23-sim}
Y.~Sun, S.~Zhang, J.~Shan, L.~Zhao, X.~Wang, F.~Sun, Y.~Yang, and B.~Fang, ``Simulation of vision-based tactile sensors with efficiency-tunable rendering,'' in \emph{2023 IEEE International Conference on Robotics and Biomimetics (ROBIO)}, 2023, pp. 1--6.

\bibitem{14-gel}
R.~Li, R.~Platt, W.~Yuan, A.~ten Pas, N.~Roscup, M.~A. Srinivasan, and E.~Adelson, ``Localization and manipulation of small parts using gelsight tactile sensing,'' in \emph{2014 IEEE/RSJ International Conference on Intelligent Robots and Systems}, 2014, pp. 3988--3993.

\bibitem{18-vitac}
S.~Luo, W.~Yuan, E.~Adelson, A.~G. Cohn, and R.~Fuentes, ``Vitac: Feature sharing between vision and tactile sensing for cloth texture recognition,'' in \emph{2018 IEEE International Conference on Robotics and Automation (ICRA)}, 2018, pp. 2722--2727.

\bibitem{19-cross}
J.-T. Lee, D.~Bollegala, and S.~Luo, ``“touching to see” and “seeing to feel”: Robotic cross-modal sensory data generation for visual-tactile perception,'' in \emph{2019 International Conference on Robotics and Automation (ICRA)}, 2019, pp. 4276--4282.

\bibitem{22-Taxim}
Z.~Si and W.~Yuan, ``Taxim: An example-based simulation model for gelsight tactile sensors,'' \emph{IEEE Robotics and Automation Letters}, vol.~7, no.~2, pp. 2361--2368, 2022.

\bibitem{22-TACTO}
S.~Wang, M.~Lambeta, P.-W. Chou, and R.~Calandra, ``Tacto: A fast, flexible, and open-source simulator for high-resolution vision-based tactile sensors,'' \emph{IEEE Robotics and Automation Letters}, vol.~7, no.~2, pp. 3930--3937, 2022.

\bibitem{24-difftactile}
Z.~Si, G.~Zhang, Q.~Ben, B.~Romero, Z.~Xian, C.~Liu, and C.~Gan, ``Difftactile: A physics-based differentiable tactile simulator for contact-rich robotic manipulation,'' 2024.

\bibitem{93-FEM}
S.~Ricker and R.~Ellis, ``2-d finite-element models of tactile sensors,'' in \emph{[1993] Proceedings IEEE International Conference on Robotics and Automation}, 1993, pp. 941--947 vol.1.

\bibitem{19-FEM}
C.~Sferrazza, A.~Wahlsten, C.~Trueeb, and R.~D’Andrea, ``Ground truth force distribution for learning-based tactile sensing: A finite element approach,'' \emph{IEEE Access}, vol.~7, pp. 173\,438--173\,449, 2019.

\bibitem{93-effect}
N.-S. Lee and K.-J. Bathe, ``Effects of element distortions on the performance of isoparametric elements,'' \emph{International Journal for numerical Methods in engineering}, vol.~36, no.~20, pp. 3553--3576, 1993.

\bibitem{13-snow-mpm}
\BIBentryALTinterwordspacing
A.~Stomakhin, C.~Schroeder, L.~Chai, J.~Teran, and A.~Selle, ``A material point method for snow simulation,'' \emph{ACM Trans. Graph.}, vol.~32, no.~4, jul 2013. [Online]. Available: \url{https://doi.org/10.1145/2461912.2461948}
\BIBentrySTDinterwordspacing

\bibitem{16-mpm}
\BIBentryALTinterwordspacing
C.~Jiang, C.~Schroeder, J.~Teran, A.~Stomakhin, and A.~Selle, ``The material point method for simulating continuum materials,'' ser. SIGGRAPH '16.\hskip 1em plus 0.5em minus 0.4em\relax New York, NY, USA: Association for Computing Machinery, 2016. [Online]. Available: \url{https://doi.org/10.1145/2897826.2927348}
\BIBentrySTDinterwordspacing

\bibitem{21-elastic-sim}
\BIBentryALTinterwordspacing
Y.~Wang, W.~Huang, B.~Fang, F.~Sun, and C.~Li, ``Elastic tactile simulation towards tactile-visual perception,'' in \emph{Proceedings of the 29th ACM International Conference on Multimedia}, ser. MM '21.\hskip 1em plus 0.5em minus 0.4em\relax New York, NY, USA: Association for Computing Machinery, 2021, p. 2690–2698. [Online]. Available: \url{https://doi.org/10.1145/3474085.3475414}
\BIBentrySTDinterwordspacing

\bibitem{Cycles}
\BIBentryALTinterwordspacing
``Cycles: open source production rendering.'' [Online]. Available: \url{https://www.cycles-renderer.org/features/}
\BIBentrySTDinterwordspacing

\bibitem{17-Blender-1}
\BIBentryALTinterwordspacing
M.~Jaros, L.~Riha, T.~Karasek, P.~Strakos, and D.~Krpelik, ``Rendering in blender cycles using mpi and intel® xeon phi™,'' in \emph{Proceedings of the 2017 International Conference on Computer Graphics and Digital Image Processing}, ser. CGDIP '17.\hskip 1em plus 0.5em minus 0.4em\relax New York, NY, USA: Association for Computing Machinery, 2017. [Online]. Available: \url{https://doi.org/10.1145/3110224.3110236}
\BIBentrySTDinterwordspacing

\bibitem{21-Blender-2}
\BIBentryALTinterwordspacing
F.~Xie, P.~Mishchuk, and W.~Hunt, ``Real time cluster path tracing,'' in \emph{SIGGRAPH Asia 2021 Technical Communications}, ser. SA '21.\hskip 1em plus 0.5em minus 0.4em\relax New York, NY, USA: Association for Computing Machinery, 2021. [Online]. Available: \url{https://doi.org/10.1145/3478512.3488605}
\BIBentrySTDinterwordspacing

\bibitem{13-Blender-3}
T.~M. Takala, M.~Mäkäräinen, and P.~Hämäläinen, ``Immersive 3d modeling with blender and off-the-shelf hardware,'' in \emph{2013 IEEE Symposium on 3D User Interfaces (3DUI)}, 2013, pp. 191--192.

\bibitem{24-thu-sensor}
S.~Zhang, Y.~Yang, F.~Sun, L.~Bao, J.~Shan, Y.~Gao, and B.~Fang, ``A compact visuo-tactile robotic skin for micron-level tactile perception,'' \emph{IEEE Sensors Journal}, pp. 1--1, 2024.

\end{thebibliography}
\vfill
\end{document}